%% file: root.tex
\documentclass[letterpaper, 10 pt, conference]{ieeeconf}  
\usepackage[utf8]{inputenc}

\IEEEoverridecommandlockouts                              

\usepackage[export]{adjustbox}
\usepackage{graphicx}
\usepackage{mathtools}
\usepackage{xfrac}
\usepackage[english]{babel}
\usepackage[T1]{fontenc}
\usepackage{amsmath,amssymb,bbm} 
\usepackage{array}
\usepackage{mathrsfs}
\usepackage{cite}
\usepackage{multirow}
\usepackage{amsmath}
\usepackage{subfig}
\usepackage{multirow}
\usepackage{color}
\usepackage{url}
\usepackage[table]{xcolor}
\newcommand{\squeezeup}{\vspace{-2.5mm}}
\newcommand*\rot{\rotatebox{90}}

\title{\LARGE \bf
    A Multimodal Classifier Generative Adversarial Network\\ for Carry and Place Tasks from Ambiguous Language Instructions
}

\author{Aly Magassouba, Komei Sugiura and Hisashi Kawai
 \thanks{National Institute of Information and Communication Technology, 3-5 Hikaridai, Seika, Soraku, Kyoto 619-0289, Japan  {\tt\small aly.magassouba@nict.go.jp}}%
}

\newcommand*\Update{\color{black}}
\newcommand*\Done{\color{black}}
\begin{document}

\maketitle
\thispagestyle{empty}
\pagestyle{empty}
\graphicspath{{figures/}} 
\input{tex/abstract}
\input{tex/introduction}

\input{tex/related}
\input{tex/pb_statement}

\input{tex/method}

\input{tex/label}

\input{tex/experimentation}

\input{tex/conclusion}
\input{tex/appendix}






\bibliographystyle{IEEEbib}
\bibliography{strings,bib/bibthese}

\end{document}

%% file: tex/abstract.tex
\begin{abstract}
This paper focuses on a multimodal language understanding method for carry-and-place tasks with domestic service robots. 
We address the case of ambiguous instructions, that is, when the target area is not specified. For instance ``put away the milk and cereal'' is a natural instruction where there is ambiguity regarding the target area, considering environments in daily life. Conventionally, this instruction can be disambiguated from a dialogue system, but at the cost of time and cumbersome interaction. Instead, we propose a multimodal approach, in which the instructions are disambiguated using the robot's state and environment context. We develop the Multi-Modal Classifier Generative Adversarial Network (MMC-GAN) to predict the likelihood of different target areas considering the robot's physical limitation and  the target clutter. 
Our approach, MMC-GAN, significantly improves accuracy compared with baseline methods that use instructions only or simple deep neural networks.
\end{abstract}


%% file: tex/introduction.tex
\section{Introduction}

With the growth of an aging population, improving the quality of life of this  segment of the population is a major issue for society.
Robots represent a credible solution providing support to elderly and/or disabled people.
Recently, domestic service robots (DSRs) hardware and software are being standardized and many studies have been conducted  \cite{piyathilaka2015human, smarr2014domestic,iocchi2015robocup}. However, in most DSRs, the communication ability is limited. However, for communicative DSRs, it is crucial to be able to interpret user commands for object manipulation tasks. These commands are naturally generated from language.

In this context, our work focuses on language understanding for ``carry-and-place'' tasks. We define carry-and-place tasks as those in which the robot moves an object from an initial area to a target area.  The main challenge of using natural language is that linguistic information may be ambiguous or insufficient: in our configuration, we assume that the target area for placing the object is missing  or insufficiently specified in the instruction, {\it e.g.}, ``Put away the milk and cereal.'' 

A simple approach consists of asking the user for the missing information ({\it e.g.}, Robocup@Home \cite{iocchi2015robocup}). Nevertheless, it sometimes takes more than one minute to disambiguate the instruction, which is cumbersome. More advanced approaches \cite{kollar2013learning, gemignani2015language} rely on linguistic knowledge with the development of dialog to fill in the missing slots in the language understanding model. Unfortunately, these methods can also be time-consuming.


 \begin{figure}[tp]
   \centering
      \includegraphics[scale=0.28]{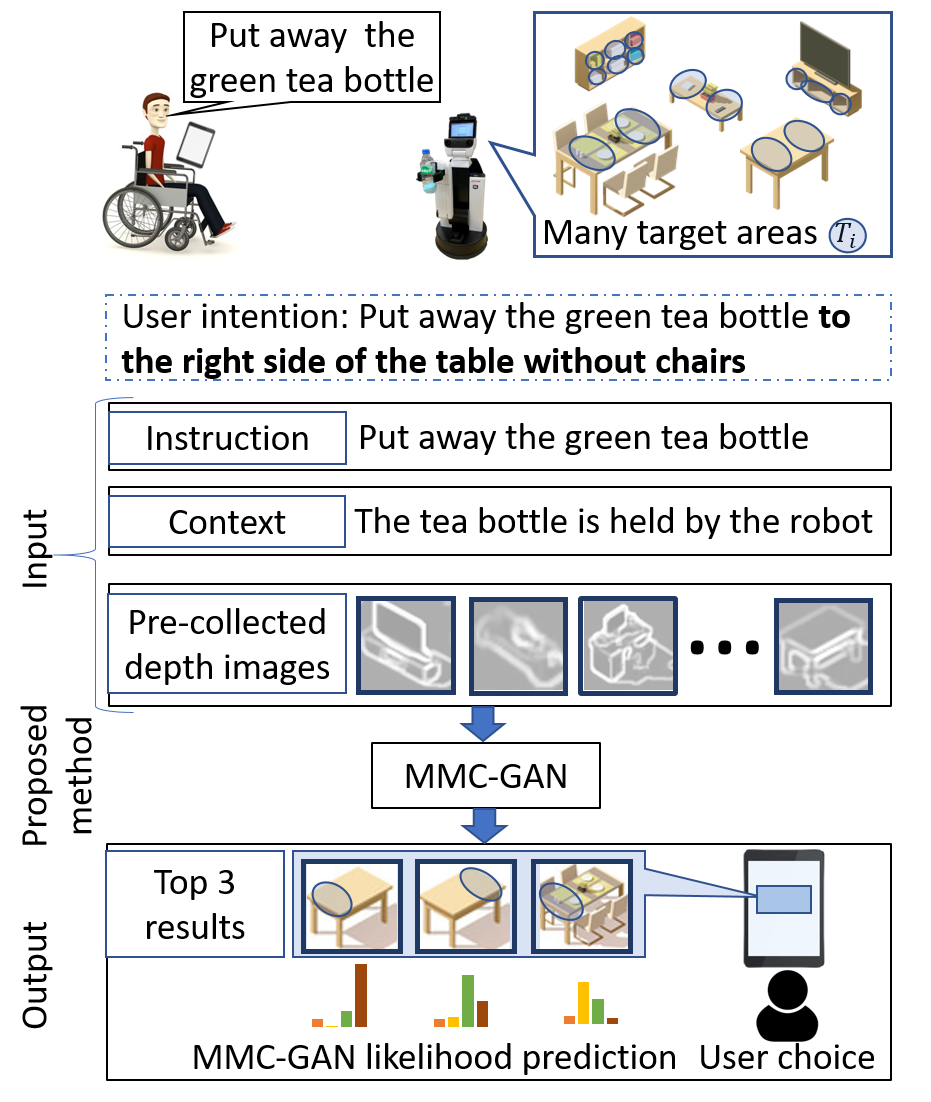}
      \caption{\small Architecture of our method to solve ambiguous instructions based on MultiModal Classifier GAN (MMC-GAN).}
   \label{fig:method}
   \squeezeup
   \squeezeup
   \squeezeup
 \end{figure}

In this work, we propose a method that allows us to predict the likelihood of target areas for ambiguous instructions by taking into account the physical ability of the robot as well as the available space. We solve this problem by introducing a generative adversarial nets (GAN) \cite{goodfellow2014generative} classifier that is able to address multimodal inputs called Multi-Modal Classifier GAN (MMC-GAN). 

Recently several studies have employed GAN-based approaches for classification tasks \cite{springenberg2015unsupervised, sugiura2018grounded}: however they only use a single modality.  
In contrast, our approach is based on  linguistic (user instructions and scene context) as well as visual inputs related to the different target area candidates. Using a latent space representation of these inputs, MMC-GAN can address both modalities through a unified framework. As a result, our classification method accuracy exceeds $80\%$. A demonstration video is available at this URL\footnote{\protect\url{https://youtu.be/_YQuziz4eGY}}.

The key contributions of this paper can be summarized as follows:
\begin{itemize}
\item[$\bullet$] A multi-modal GAN classifier based on the user instruction, task context, and depth image of a given target area is introduced in Section \ref{sec:prop}. 
\item[$\bullet$] Inspired by the wide literature on GAN, we propose and evaluate several variations of MMC-GAN  in Section \ref{sec:exp}.
\end{itemize}

%% file: tex/related.tex
\section{Related work}

Inferring a user's intention does not only  rely on linguistic inputs but also on proprioceptive and contextual knowledge. Several studies in the robotic community focus on mapping instructions to the environment context. In \cite{misra2016tell} manipulation tasks are addressed based on cloud data, while in \cite{tellex2011understanding} navigation and path planning tasks are addressed. 

Like many pick-and-place approaches, we are interested in placing tasks in daily life environments. However, most of these approaches focus either on the grasping and manipulation part \cite{jiang2012learning} or on the the method of placing an object. Assumptions are made that the target areas already specified and available. This is the case in  \cite{abdo2015robot} where  the authors proposed a solution based on user preferences for placing objects on shelves, and  in \cite{schuster2010perceiving} where objects are placed on the uncluttered parts of flat surfaces using image segmentation. More importantly, these studies do not focus on the instruction understanding of the robot.

In contrast, we focus on determining suitable target areas for placing an object in a realistic environment when the target area is undefined in the user instruction. In this way, we think that our work complements these studies.  

Recently, GAN and all its variation have spurred the field of image reconstruction and enhancement \cite{ledig2016photo, denton2015deep}.  Interestingly, GAN-based approaches have also been used to address classification problems \cite{springenberg2015unsupervised, sugiura2018grounded, odena2016conditional}.  The latter studies proposed to improve the  classification task by exploiting the data augmentation property of a GAN. Our work, extending GAN by considering multimodal inputs, is inspired by these methods.


%% file: tex/pb_statement.tex
\section{Problem Statement}\label{sec:prob}

\subsection{Task Description}
The target task of this study is the understanding of ambiguous language instructions in carry-and-place tasks. Typical instructions follow a pattern that can be described by the sentence  \textbf{``Put away an object ($O$) (in a target area $T_i$)''}. As an appropriate response to this instruction, a robot should be able to predict a suitable target area $T_i$, where $T_i$ is not explicitly or fully specified in the instruction.

We assume the following inputs and output:
\Update
\begin{itemize}
      \item[$\bullet$]{\bf Inputs}:  Linguistic instruction, linguistic context and pre-collected camera images of candidate target areas.
      \item[$\bullet$]{\bf Output}: Likelihood of the target areas.
\end{itemize}
The likelihood refers to the probability that the robot should place the designated object in a given target area for four output classes. The target areas are afterwards ranked by simple binarization from four classes to two classes (sum of the two best class and two worst class probabilities). Hence, this is not a multi-class evaluation of pieces of furniture but a multi-class evaluation  to determine the suitability of the target areas, described as a four-class problem and detailed in Section \ref{sec:label}.
\Done

The first type of solution to solve this problem is a dialogue-based approach \cite{kollar2013learning,johnson2011enhanced}  to recover the missing information through explicit instructions from the user. However, this solution can be cumbersome for the user, when the robot starts a dialog for each task it must perform.  Because our focus is on how to reduce the cumbersomeness of the interaction, no additional dialogue is allowed. Furthermore, because only the top $n$ candidates are shown for usability, a rank based on the target likelihood should be an appropriate output.

We assume that this task depends on the space available and the robot's physical ability. The task environment is composed of several pieces of furniture $F$ such as shelves, bookshelves, tables, desks, or other items and may contain one or several target areas $T_i$ each.  Each target  $T_i$ may also be cluttered with obstacles.

Moreover, some instructions might also be ambiguous from the context of the task. For instance, this is stressed by the instruction, \textbf{``Move the milk box ($O$) on the table ($T_i$)''}. Depending on  whether the robot is carrying $O$ or not, this instruction can lead to two distinct actions: putting down $O$ on the table  or picking up $O$ on the table and placing it somewhere else.

For visual inputs, while RGB data is mainly exploited in the computer vision community, we assume a set of visual inputs only composed of depth information D. In our problem, the depth image provides sufficient information, about target areas, to predict the likelihood of a given $T_i$ as it is shown in Section \ref{sec:exp}.
\begin{figure}[tp]
  \centering
    \subfloat[HSR]{\label{fig:hsr}\includegraphics[width= 3.5cm, height=4 cm]{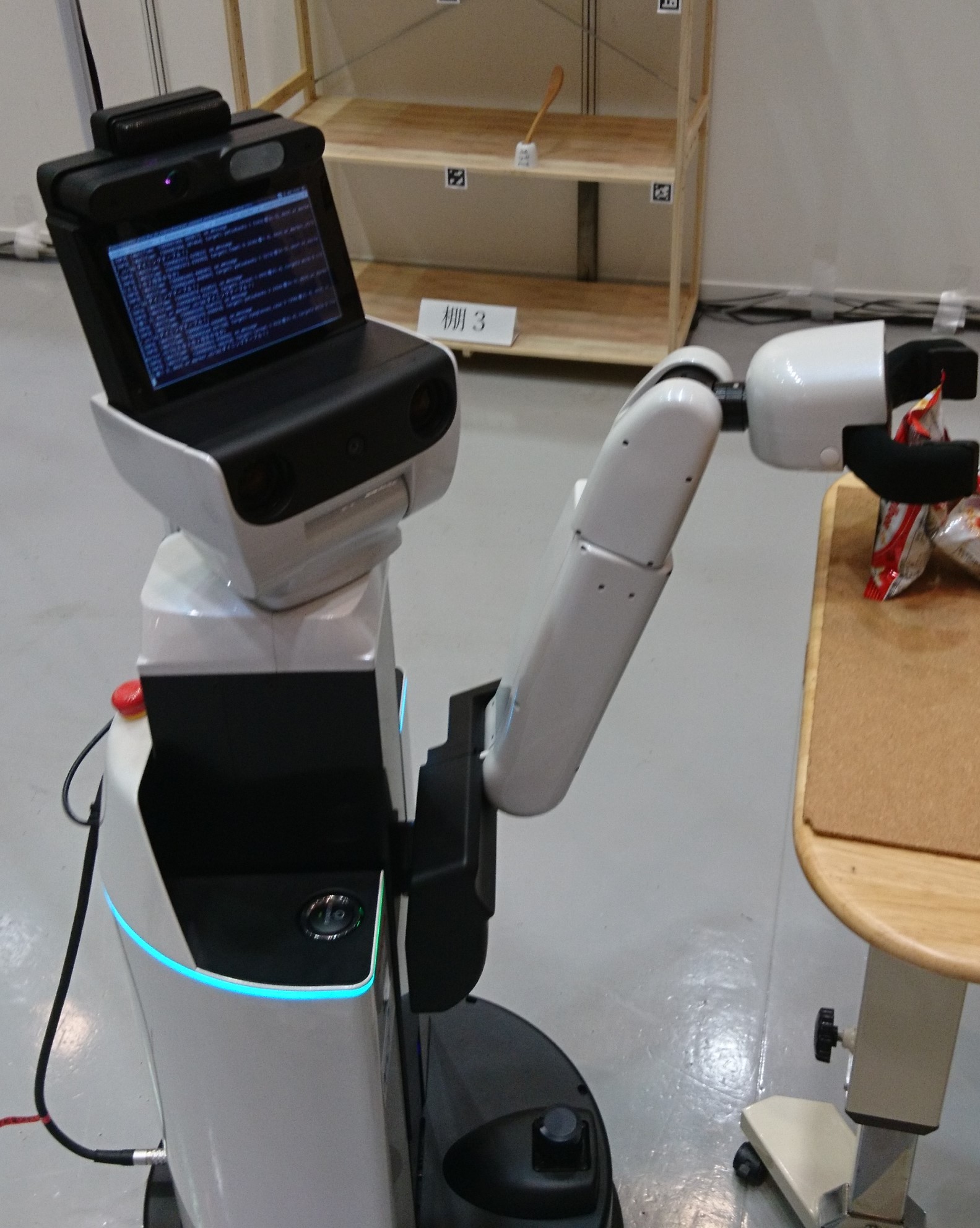}} \enskip
    \subfloat[NICT case]{\label{fig:nict_case}\includegraphics[width= 3.cm, height=4 cm] {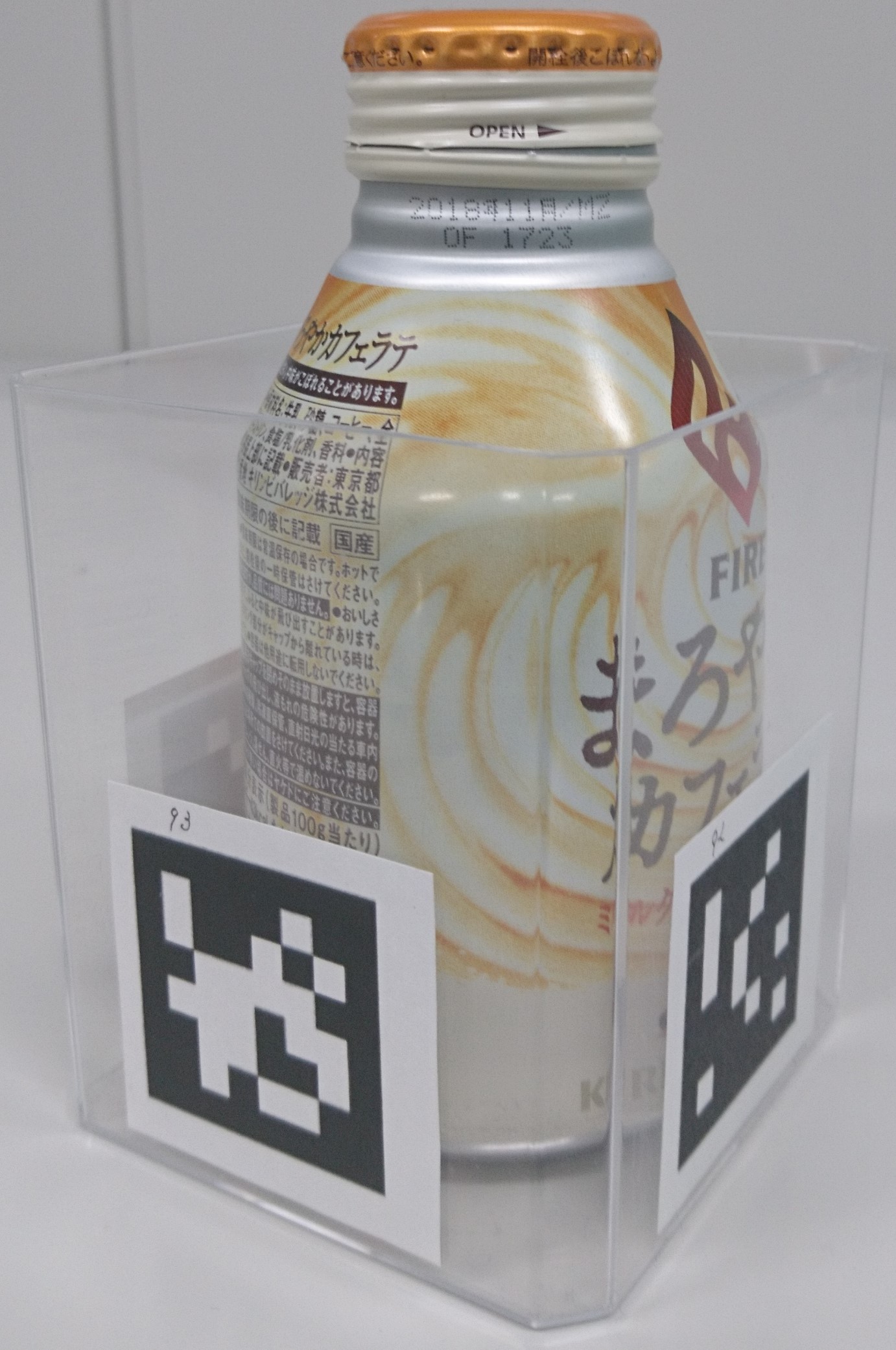}}
     \caption{\small NICT cases allow  HSR to manipulate different types of object easily. }
     \squeezeup
     \squeezeup
\end{figure}

\subsection{Hardware Assumptions}
We consider a standardized robotic platform because a target area likelihood  also depends on the robot's limitations. For this work, we use HSR (Human Support Robot),  a service robot developed by Toyota endowed with object manipulation capability (see Fig. \ref{fig:hsr}). Since 2017, this robot has been used as the domestic standard platform of RoboCup@Home competitions \cite{iocchi2015robocup}.

Moreover, because we are not dealing with the grasping problem, we introduce  ``Nothing-is Corresponding To-the-marker (NICT)'' cases (see Fig. \ref{fig:nict_case}) which are the containers of the movable objects $O$ in the scene. These cases simplify the grasping task, while not being particularly inconvenient for the user. In this case, the robot has to manipulate rigid bodies with a known shape, independently of the object type or consistency inside.


%% file: tex/method.tex
   \section{LAtent Classifier-GAN}\label{sec:def}
   To solve the target task, we extend the LAtent Classifier GAN (LAC-GAN) proposed in \cite{sugiura2018grounded}. Our choice is motivated by the data augmentation property of the GAN framework that is particularly interesting in robotics. In contrast to computer vision or speech recognition, a large-scale dataset cannot be collected due to the  diversity of environments and robotic setups. Hence, the data augmentation property of GAN allows us to design/use complex models without overfitting to the dataset.
   Moreover, from LAC-GAN's initial structure, we can exploit latent space features that are more suitable for processing multimodal inputs.
  
   \subsection{ Generative Adversarial Nets }
The GAN framework  was initially proposed in \cite{goodfellow2014generative} as a generative framework and is composed of two networks, a Discriminator $D$ and a Generator $G$, which compete with each other. $G$ creates fake data by mimicking the real data distribution while $D$ discriminates the real data from the fake data. By mutual enhancement, $G$ is trained to create more realistic fake data, while the discrimination ability of $D$ improves.
   More formally, $G$ is a network with a multi-dimensional random input ${\bf z}$ that outputs the data ${\bf x}_{fake}$:  
   \begin{equation}\label{equ:G_out}
   {\bf x}_{fake}= G({\bf z}).
   \end{equation}
  The input ${\bf z}$ is generally sampled from a standard normal distribution and its details for our case study are given in Section \ref{sec:exp}.

   In contrast, $D$ whose task is to discriminate the real data ${\bf x}_{real}$ from ${\bf x}_{fake}$, is alternately input with ${\bf x}={\bf x}_{real}$ or ${\bf x}={\bf x}_{fake} $ from a source $S \in \{real,fake\}$. $D$'s output is then given by
   \begin{equation}\label{equ:D_out}
   p_D(S=real|{\bf x})=D({\bf x}).
   \end{equation}
   The cost functions of $G$ and $D$, respectively $J_G$ and $J_D$, are defined as follows: 
   \begin{equation}\label{equ:J_S}
   	J_S=-\frac{1}{2} \mathbb{E}_{{\bf x}_{real}} \log D ({\bf x}_{real}) - \frac{1}{2} \mathbb{E}_{\bf z} \log(1-D({\bf x}_{fake})),
   \end{equation} 
   leading to $J_D=J_S$ and $J_G=-J_S$.

   \subsection{Classification using LAC-GAN }
In contrast to GAN, LAC-GAN is designed for classification tasks, in which the generated samples from $G$ are used for data augmentation. However, unlike the other GAN-based classification methods based on raw inputs \cite{odena2016conditional} ({\it e.g.}, images or text), LAC-GAN uses a third component, the extractor $E$ network, to process latent space features in GAN. Actually, $G$  does not necessarily produce a raw data representation if the task is not generation but classification. 

\begin{figure}[tp]
   \centering
       \includegraphics[scale=0.4]{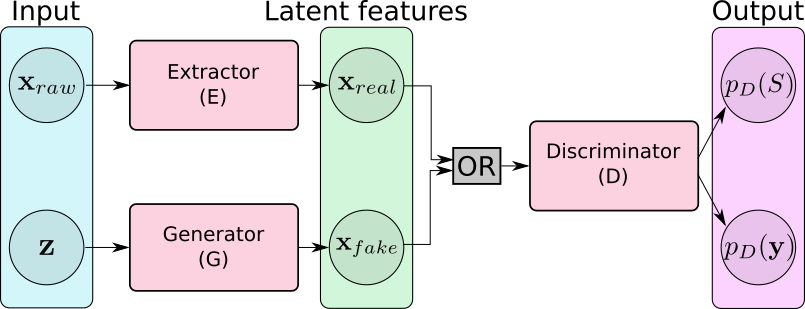}
       \caption{\small LAC-GAN architecture involving three components that are the Extractor, Generator and Discriminator.}
   \label{fig:lacgan}
   \squeezeup
   \squeezeup
   \end{figure}

   As a result, the input of $E$ is the raw data ${\bf x}_{raw}$ while the latent space feature ${\bf x}_{real}$ is output and injected into $D$.  Hence, in addition to \eqref{equ:D_out},  $D$ has a second output $ p_D({\bf y})$ which is the predicted category output. According to this new structure, the cost function of $D$ is modified as follows:
   \begin{equation}\label{equ:J_D}
   	J_D=J_S+ \lambda J_C,
   \end{equation} 
   where $\lambda$ is a weighting parameter and $J_C$ is a cross-entropy loss function: 
   \begin{equation}\label{equ:J_C}
   	J_C=-\sum_n \sum_{j} {y^{*}_n}_j \log p_D({y_n}_j),
   \end{equation}
   where ${ y^{*}_n}_j$ denotes the label of the $j$-th dimension of the  $n$-th sample.
   Note that $J_C$ is also the cost function of $E$ by replacing $p_D({ y_n}_j)$ with $p_E({y_n}_j)$. The global framework of LAC-GAN is illustrated in Fig. \ref{fig:lacgan}.

\Update
   Such a structure implies that $E$ is trained considering the category labels ${\bf y}^{*}$. The training of LAC-GAN is then divided into two phases. First, $E$ is trained knowing the labels ${\bf y}^{*}$ with the cost function \eqref{equ:J_C}. The parameters of $E$ are detailed in Section \ref{sec:prop}. Subsequently, when the set ${\bf x}_{real}$ is extracted ({\it i.e.}, $E$ training is finished ), $D$ and $G$ are alternately trained to improve the classification prediction $p_D({\bf y})$.
\Done

  \begin{figure*}[tp]
   \centering
       \includegraphics[scale=0.35]{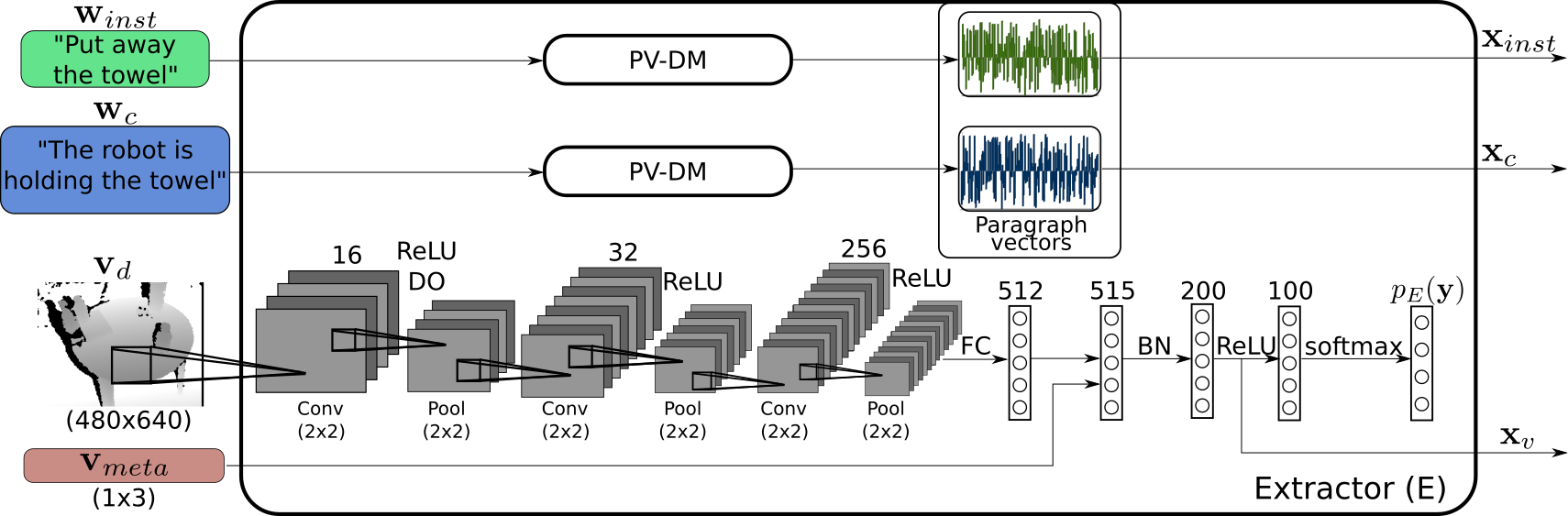}
      \caption{\small MMC-GAN addresses multimodal inputs through a CNN for visual input and a model of paragraph vector (PV-DM) for linguistic inputs. The number of nodes are shown above each layers. Both convolutional (Conv) and pooling (Pool) layers use ($2$x$2$)-sized filters. A fully connected layer (FC) is used after the convolutional layers.  Batch normalization (BN) and dropout (DO) operations are applied on the output of the first and fifth layers. } \label{fig:mmcgan}
 \squeezeup
 \squeezeup
 \end{figure*}

  \section{Proposed Method: Multimodal Classifier GAN} \label{sec:prop}

  In \cite{sugiura2018grounded}, the Extractor $E$ was defined as a fully-connected feedforward DNN. However, this structure is not optimal for multimodal inputs including visual features. The structure of $E$ should hence be modified for our study. 
 \subsection{Multimodal inputs}
We propose MMC-GAN which extends the LAC-GAN architecture to deal with multimodal inputs, namely linguistic and visual inputs.  This emphasizes another key advantage of using latent space features: with a similar representation, different inputs can be processed in the same network. Indeed, usually visual and linguistic inputs are different in dimension, which makes it difficult to generate from a unified  Generator $G$ structure. Unlike MMC-GAN, a typical GAN classifier would require us to develop a separate generator for each type of input and develop a method to merge $G$ results.
 
  In contrast to LAC-GAN, the Extractor $E$ of MMC-GAN, illustrated in Fig. \ref{fig:mmcgan}, is  composed of a convolutional neural network (CNN) and two paragraph vector models. The CNN is used to process the visual inputs, while linguistic inputs are processed via  the of a paragraph vector distributed memory (PV-DM) model  \cite{le2014distributed}. $E$'s input, ${\bf x}_{raw}$, is then defined as a set of inputs
   \begin{equation}\label{equ:E_in}
   	 {\bf x}_{raw}=\{{\bf w}_{inst}, {\bf w}_{c},{\bf v}_{d},{\bf v}_{meta}\}
   \end{equation} 
   where ${\bf w}_{inst}$ and ${\bf w}_{c}$ are the linguistic features and ${\bf v}_{d}$ and ${\bf v}_{meta}$ correspond to visual features. 
   In detail, ${\bf w}_{inst}$ and ${\bf w}_{c}$ correspond to the instruction and context sentences respectively.
   For instance a typical carry-and-place task is characterized by
   \begin{equation*}
   \left\lbrace
 \begin{aligned}
       &{\bf w}_{inst}= \text{``Move the towel to the shelf.''}\\
     &{\bf w}_{c}=\text{``The robot is holding a towel.''}
 	\end{aligned} \right. 
 \end{equation*}

    Here, ${\bf w}_{inst}$ and ${\bf w}_{c}$ are two word sequences using a one-hot vector representation that are processed through the PV-DM. As output, $200$-dimensional paragraph vectors ${\bf x}_{instr}$ and ${\bf x}_{c}$, that are also latent space features, are created from respectively from ${\bf w}_{inst}$ and ${\bf w}_{c}$. Both ${\bf w}_{inst}$ and ${\bf w}_{c}$ used for training the MMC-GAN model are more thoroughly described in Section \ref{sec:label}.
 
  For the visual inputs, ${\bf v}_{d}$ is the depth image of a target area. Although other inputs could be used ({\it e.g.}, RGB), we limited our method to depth data that is sufficient to characterize a given target area. In addition, ${\bf v}_{meta}$ describes the situation of each ${\bf v}_{d}$, that is, the candidate area height, the robot camera height and angle. 

   These inputs are processed in a CNN composed of seven layers in which  ${\bf v}_{d}$ ($640\text{x}480$ pixels) and ${\bf v}_{meta}$ ($1$x$3$) are transformed into a $200$-dimensional latent space feature ${\bf x}_{v}$. ${\bf x}_v$ is extracted from the penultimate layer ($N=6$) similarly to bottleneck network structures. The first three convolutional (Conv) layers process the depth image while a concatenation operation is performed on the fifth layer to input the metadata. We use ReLU for the activation functions. Dropout is applied to the first layer and batch normalization \cite{ioffe2015batch} is applied to the fifth layer. The cost function of the CNN, $J_{v}$ is defined as follows:
     \begin{equation}\label{equ:J_cnn}
   	 J_{v}=J_C
   \end{equation}

 As a result, the output ${\bf x}_{real}$ of $E$ is  defined as
   \begin{equation}\label{equ:E_out}
   	 {\bf x}_{real}=\{  {\bf x}_{instr},{\bf x}_{c}, {\bf x}_{v}\}.
   \end{equation} 
   where ${\bf x}_{real}$ is a $600$-dimensional vector representing the instruction, context  and a given target area.  
 
 \Update
 Note that similarly to the LAC-GAN structure, $E$ is trained beforehand to extract ${\bf x}_{real}$. Only when ${\bf x}_{real}$ is extracted, can the training of Generator $G$ and Discriminator $D$ be performed.
  \Done 
  
\subsection{Generator Architecture}
Several variations of MMC-GAN are considered in this work. These variations are based on different architectures of the Generator $G$. In addition to common GAN architecture, conditional GAN (CGAN) \cite{mirza2014conditional} and Wasserstein GAN (WGAN) \cite{arjovsky2017wasserstein} are used.

\subsubsection{CGAN}
In the CGAN architecture, the network is conditioned by ${\bf c}$, that corresponds to the category distribution, in our case. In this way, $G$ generates a feature ${\bf x}_{fake}$ following the categories given in $\bf c$, which is particularly relevant in classification problems. Indeed, in the initial GAN method, there are no constraints on the class of the data generated by $G$ which makes the training of $D$ more complex when ${\bf x}_{fake}$ and $ {\bf x}_{real}$ belong to different classes. Hence this architecture simplifies the GAN training process by matching the class of  ${\bf x}_{fake}$ to $ {\bf x}_{real}$. 

Nonetheless,  our considered CGAN architecture is different from the original one proposed in \cite{mirza2014conditional}. In our classification problem, only $G$ is conditioned by $\bf c$. Indeed $D$ which also outputs $p_D({\bf y})$ should not input $\bf y$ or $\bf c$. As a result, $G$ is modified as follows:  

\begin{equation}\label{equ:G_out2}
   {\bf x}_{fake}= G({\bf z}, {\bf c}).
\end{equation}

\subsubsection{WGAN }
 WGAN corresponds to a different training method for GAN. The aim of WGAN is to improve the stability of the  model's learning as well as to avoid mode collapse. Actually, GAN networks are known to be slow and unstable notably because of the vanishing gradient problem: the initial loss function \eqref{equ:J_S}  falls to  zero and  the training becoming slow because of a  nearly null gradient.

To solve this problem, WGAN adopts a different loss function derived from the  Wasserstein distance, which is a measure of two distributions that quantifies the cost of matching the first distribution to the second one. Considering our features ${\bf x}_{real}$ and  ${\bf x}_{fake}$, the Wasserstein distance becomes
\begin{equation}\label{equ:wass}
\sum_{{\bf x}_{fake}, {\bf x}_{real}} \gamma({\bf x}_{fake}, {\bf x}_{real}) \| {\bf x}_{fake}-{\bf x}_{real} \| 
\end{equation}
where  $\gamma({\bf x}_{fake}, {\bf x}_{real})$ represents the cost for matching  the two distributions.
From this metric, a new loss function for both $D$ and $G$ networks is defined as follows:
\begin{equation}\label{equ:J_S_wass}
J_S=- \frac{1}{2} \mathbb{E}_{{\bf x}_{real}} D ({\bf x}_{real}) + \frac{1}{2} \mathbb{E}_{\bf z} D({\bf x}_{fake})
 \end{equation}

%% file: tex/label.tex
 \section{carry and place multi-modal data set}\label{sec:label}
 In this section, we describe the carry-and-place multi-modal data set used for training and evaluating MMC-GAN. To the best of our knowledge such a data set is unavailable in the literature and has consequently been  built specially for this work.
\subsection{Overview of the Data Set Construction}\label{sec:data_constr}

The procedure given below describes the different steps used to build the carry-and-place multi-modal data set. Each of these steps will be detailed in the following sections.
\begin{itemize}
\item[$1$.] Setup an area with everyday objects.
\item[$2$.] Record the scene with a camera. \Update
\item[$3$.] Generate a pseudo-instruction sentence from randomly selected objects and pieces of furniture. 
\item[$4$.] Generate a pseudo-context sentence from randomly selected objects and pieces of furniture.
\item[$5$.] Instruct a labeler to rewrite a natural sentence based on $3$ and $4$, according to the scene.
\item[$6$.] Instruct a labeler to label the samples according to $6$ and $2$ .
\end{itemize} 
\Done
 \subsection{Depth Image Inputs} \label{sec:depth}
\begin{figure}[b]
   \centering
      \includegraphics[scale=0.45]{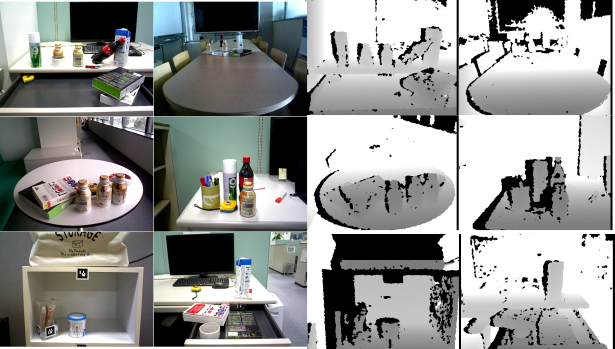}
      \caption{\small Samples of the $1282$ depth images (right) ${\bf v}_d$ and their corresponding RGB images (left). Only the depth data is used.}
   \label{fig:data}
   \squeezeup
 \end{figure}

Each linguistic input namely ${\bf w}_{inst}$ and ${\bf w}_c$ should be associated with a couple of visual  features $\{{\bf v}_d, {\bf v}_{meta}\}$  obtained as follows.  To characterize the variability of daily life environments, different types of tables, drawers, shelves or desks, from various areas of our office building, were used to build the set ${\bf T}$. The dataset of visual inputs ${\bf v}_d$ is then based on the depth image of a candidate area $T_i$ ($T_i \in {\bf T}$) with different levels of clutter and layout variations as illustrated in Fig. \ref{fig:data}.

 The depth images were collected from an \textit{Asus Xtion Pro live} depth sensor, which is identical to the one equipped on HSR. As summarized in Table \ref{tab:data_coll},  $1282$ valid images could be collected from $37$ target areas belonging to four drawers, four tables, four desks and two shelves. For each target area $T_i$, several configurations were recorded. We used $19$ objects as obstacles, and varied the obstacle layout for each image recorded. 
 \begin{table}[t]

\normalsize
\caption{\small Data collection statistics }
\label{tab:data_coll}
\centering
\begin{tabular}{lccc}
\hline
$\#$ &Instance &Target areas &Images\\
\hline
Table &4&18&321 \\
  
Drawer &4 &10&336\\

Desk&4&8&425 \\

Shelf&2&5&200 \\
\hline
$\sum$ &{\bf 14}& {\bf 37}& {\bf 1282}\\
\hline
\end{tabular}
    \squeezeup
   \squeezeup

\end{table}
 Note that the depth images recorded by this type of sensor are particularly noisy. Blobs related to missing information (represented as black pixels in Fig. \ref{fig:data}) deteriorate the quality of the image and add complexity to the classification task. 
 Eventually, ${\bf v}_{d}$ was coupled with the corresponding ${\bf v}_{meta}$. The input ${\bf v}_{meta}$ enables us to differentiate the observation points of the robot ({\it i.e.}, image perspective).  

\subsection{Instructions and contexts}
\begin{figure}[tp]
   \centering
      \includegraphics[scale=0.4]{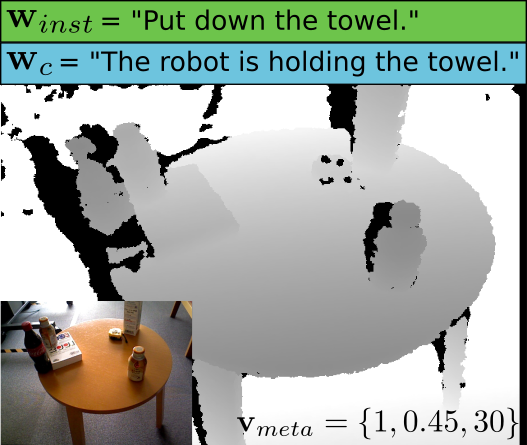}
      \caption{\small Example of multimodal input composed of a depth image ${\bf v}_d$ and the associated meta-data ${\bf v}_{meta}$ (camera height, target height, camera angle) with the instruction ${\bf w}_{inst}$ and context ${\bf w}_{c}$.}
   \label{fig:features}
   \squeezeup
   \squeezeup
  \end{figure}

\Update
This  dataset relates a linguistic input to each visual input (see Fig. \ref{fig:features}). We assume that a pseudo-sentence is randomly generated from a randomly selected target piece of furniture and/or object .
A labeler is then instructed to rewrite a natural language sentence based on a randomly selected verb phrase, a noun phrase and a preposition phrase. The labeler is also instructed to rewrite the natural sentences under two conditions so as to obtain two sets $Y$ and $N$ characterized by the instructions $\{I_Y,I_N\}$:
\begin{itemize}
    \item $I_{Y}$: A target piece of furniture $F$ should be mentioned in the instruction, e.g., ``Move the coke bottle to the kitchen and put it down on the table''
    \item $I_N$: The instruction should not contain any target piece of furniture, e.g., ``Move the coke bottle to the kitchen and put it down''
\end{itemize} 
The context sentence is obtained in a similar manner by the labeler writing a natural sentence. The context sentence ${\bf w}_{c}$ refers to the situation of the robot with respect to carried object $O$. Example context sentences are : 
``$O$ is held by the robot'',  ``$O$ is already on $F$'',  ``The robot is holding $O$'' and  ``$O$ can be found on $F$''. 
Note that we considered the case where instruction/context sentences contain a target piece of furniture $F$. 


\Done


\subsection{Labeling Strategy}\label{sec:labeling}
The samples were manually labeled into categories by a robotic expert. The categories are based on the physical abilities of HSR and the scene context. The labeling has been performed by a single labeler;  having multiple labelers will be considered in future work. 

\Update
The labeling strategy should emphasize  several concepts. The accessibility to an area of action: areas that would be too far, too high (approx. above $0.8$ m) or too low (approx. below $0.6$ m) for the robot should be penalized. The scene clutter: collision with obstacles or even with the given piece of furniture edge/border should be avoided. Moreover, the scene should match with the initial instruction, in particular, when target pieces of furniture are specified.

With the knowledge of HSR's capability ({\it e.g.}, arm's reach and gripper size), for each sample ${\bf x}_{raw}$, we specify the labeling rule as follows. 
\begin{itemize}
\item[($A1$)]  The target area $T_i$ is very likely in terms of linguistic and visual information: the area is  within the natural reach of the robot and have enough space. 
\item[($A2$)]  The target area $T_i$ is likely in terms of linguistic and visual information: however the area is not within the natural reach of the robot or have few clutters. 
\item[($A3$)] The target area $T_i$  is unlikely in terms of linguistic and/or visual information: there is limited space available on $T_i$, which might lead to collisions with obstacles. 
\item[($A4$)] The target area $T_i$ is very unlikely in terms of linguistic and/or visual information: there is not enough space on $T_i$, with obstacles preventing the robot from placing the object. 
\end{itemize}  
\Done
From this labeling strategy, after removing the invalid features, our initial data set was randomly  split into training, validation and test sets.  We obtained the different sets given in Table \ref{tab:data_table}.
\Update
It should also  interesting to mention that during the labeling phase, the instructions referring to tables (resp. desks) could match with images of desks (resp. tables), depending on the evaluation of the labeler.
\Done

\begin{table}[h]
\normalsize
\caption{\small Statistics of the data set ${\bf x}_{real}$}
\label{tab:data_table}
\centering
\begin{tabular}{cccc|c }
\hline
$\#$ & Train& Valid& Test& $\sum$\\
\hline
$A1$  &158&29&25&{\bf 212} \\

$A2$&359&34&39&{\bf 432}\\

$A3$&350&26&22 &{\bf 398}  \\

$A4$&203&17&20 & {\bf 240}\\
\hline
$\sum$& {\bf 1070} ({\it 83 $\%$}) & {\bf 106} ({\it 8.5 $\%$}) & {\bf 106} ({\it 8.5 $\%$}) & \multicolumn{1}{ |c}{\bf 1282}\\
\hline
\squeezeup
\squeezeup
\squeezeup
\end{tabular}
\end{table}

%% file: tex/experimentation.tex
\section{EXPERIMENTS}\label{sec:exp}

\subsection{Setup}
The parameter settings of MMC-GAN are summarized in Table \ref{tab:param}. 
\begin{table}[b]
\squeezeup
\squeezeup
\normalsize
\centering
\caption{\small Parameter settings and structures of MMC-GAN}\label{tab:param}
\begin{tabular}{|c|c|l|}
\hline
&Opt. & Adam (Learning rate= $0.00005$, \\
&method & $\beta_1=0.5$, $\beta_2=0.9$), $\lambda$=0.2 \\
\hline
 &Batch & $64$ (E), $50$ ($G$ and $D$)\\
\hline
\hline
 Num. &GAN & $100$, $100$, $100$, $100$ \\
\cline{2-3}
nodes  &CGAN  & $100$, $100$, $100$, $100$  \\
\cline{2-3}
 (G) &WGAN & $100$, $100$, $100$, $100$  \\
\hline
\hline
&(1) &  $100$, $200$ \\
\cline{2-3}
Num.&(2) & $100$, $200$, $400$ \\
\cline{2-3}
 nodes   & (3) &  $100$, $200$, $400$, $1000$ \\
\cline{2-3}
(D) &(4) &  $100$, $200$, $400$, $1000$, $1000$\\
\hline
\end{tabular}
\end{table}

\begin{table*}[h]
\normalsize
\caption{\small Mean validation/test-set accuracy and sample standard deviation for the best model of each method considering the instruction only (I), the instruction and context only (I+C), the visual features only (V) and the instruction, context and visual features (I+C+V). These results are based on ten random experiments. (*) report partial results for configurations when all the experiments did not converge.}
\label{tab:results}
\centering
\begin{tabular}{cc|cc|cc|cc|cc}
\hline
\multicolumn{1}{c}{}&\multicolumn{1}{c|}{$[\%]$} &\multicolumn{8}{c}{Input features }\\
\hline
&GAN& \multicolumn{2}{c|}{\bf I} & \multicolumn{2}{c|}{\bf I + C}& \multicolumn{2}{c|}{\bf V}&\multicolumn{2}{c}{\bf I + C + V} \\
  \cline{3-10}
Method & type & Valid & Test& Valid & Test& Valid & Test & Valid& Test\\
\hline
\hline
Baseline &- &61.4$\pm$0.5&59.4$\pm$0.3 &60.7$\pm$1.5  &60.2$\pm$0.8& 63.3$\pm$0.5 & \bf 61.1$\pm$1.1&83.1$\pm$1.7 &82.2$\pm$2.8 \\
\hline
Ours  & GAN &59.3$\pm$1.4&57.5$^*$$\pm$3.3 &58.0$\pm$2.2 &59.5$^*$$\pm$ 3.7 &60.3$\pm$1.0 & 58.1$\pm$1.5&85.9$\pm$0.4&85.3$\pm$1.2\\
\hline
Ours & CGAN& 60.1$\pm$1.5& 56.4$^*$$\pm$3.7& 58.7$\pm$1.7&56.7$^*$$\pm$4.2 &57.0$\pm$0.8 & 58.2$\pm$1.0&86.6$\pm$0.3 &\bf 86.2$\pm$0.8 \\
\hline
Ours & WGAN &62.0$\pm$1.5  & \bf 61.8$\pm$2.1& 63.3$\pm$0.4  &\bf 62.7$\pm$2.1 &59.6$\pm$3.1 &59.7$\pm$1.9&84.1$\pm$0.4   &84.4$\pm$1.1\\ 
\hline
\end{tabular}
\squeezeup
\squeezeup
\end{table*}

\subsubsection{Extractor}
As discussed in the previous section, \Update the visual inputs  ${\bf v}_{d}$ and ${\bf v}_{meta}$ \Done are processed in the Extractor $E$ through a CNN. This CNN (see Fig. \ref{fig:mmcgan}) was already trained over $100$ epochs to obtain stable latent space features. The latent features were then extracted from the model for which the validation set accuracy had reached its maximum value. Note that $E$ is trained considering the labeled data. From the CNN output, we obtain ${\bf x}_v$ of dimension $d_v=200$. 

In parallel, the paragraph vector model was trained from a corpus of 4.72 million sentences similarly to \cite{sugiura2018grounded}. The output of the PV-DM generated ${\bf x}_{inst}$ and ${\bf x}_c$ which have respectively a dimensions $d_{inst}=200$ and $d_{c}=200$. We then obtained the features ${\bf x}_{real}$  with $d_{real}=600$.
 
\subsubsection{Generator}
For the MMC-GAN, we developed several versions of  the \Update Generator $G$ and Discriminator $D$ \Done as summarized in Table \ref{tab:param}. Generator $G$ is composed of four layers using ReLU activation functions, except for the last layer which uses a tanh activation function. Batch normalization was also applied to these layers. The random variable ${\bf z}$ is defined as ${\bf z}={\bf z}_1\sim \mathcal{N}(0,N)$ with a dimension $d_z=100$ in the GAN and  \Update WGAN (Wasserstein GAN) approaches \Done. For the \Update CGAN (Conditional GAN) \Done approach ${\bf z}= \{{\bf z}_1,{\bf c}\}$ is used and $\bf c$ is sampled from a categorical distribution.  For the latter case, the input dimension is $d_z=104$ because we  consider a 4-class problem. The output of the generator has dimension $d_{fake}=600$.

\subsubsection{Discriminator}
\Update
For $D$ we consider several structures ($(1)$ to $(4)$) for which we varied the number of layers and/or nodes. These  structures are detailed in the   Table \ref{tab:param}. \Done

\subsection{Results}
\begin{figure}[t]
  \centering
  \includegraphics[scale=0.38]{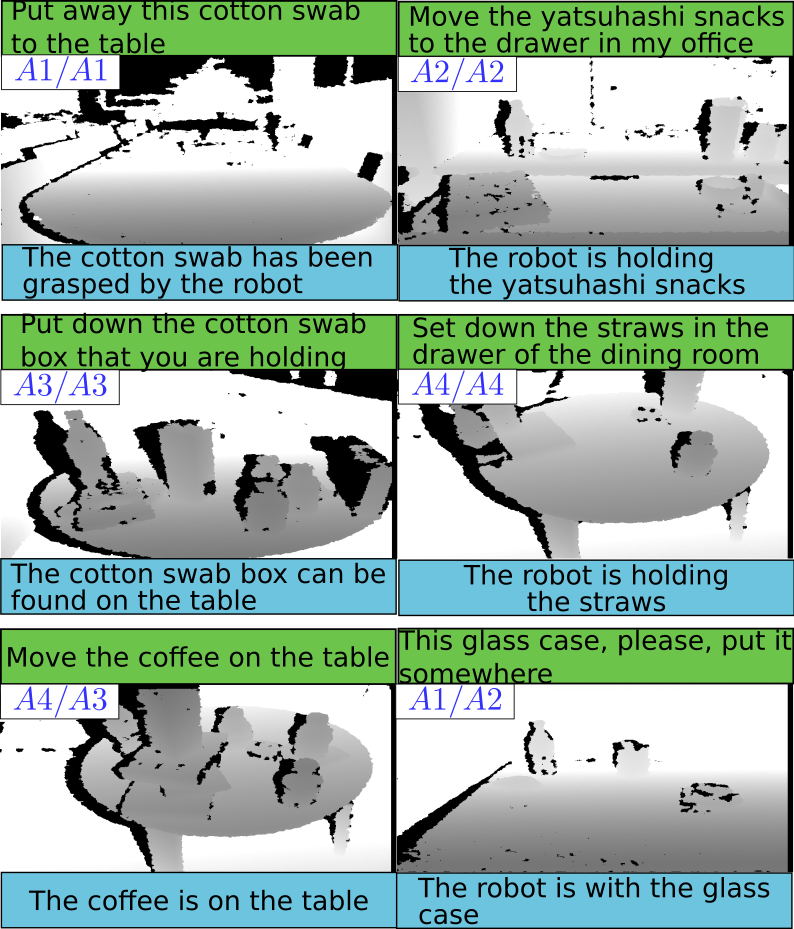} 
  \caption{\small Prediction samples (left corner true/predicted) for given instructions, contexts and depth images. The samples are mostly correctly classified except for the bottom  samples.}\label{fig:samples}
  \squeezeup
  \squeezeup
\end{figure}

Qualitative results of MMC-GAN prediction are shown in Fig. \ref{fig:samples} which illustrates typical true and false predictions. 
These errors are more thoroughly analyzed in Section \ref{sec:error_analysis}.


We evaluated the accuracy of different MMC-GAN variations in a standard way; that is the test set accuracy when the best validation accuracy is obtained. Hence, these results reflect more objectively the performance of MMC-GAN models on unseen data.

The results reported in Table \ref{tab:results} are based on the mean accuracy and sample standard deviation, over ten trials. Our methods (MMC-GANs) are compared with a baseline method (simple DNN) with the same dataset. For a fair comparison, the DNN has the same architecture as $D$ (structure $4$ in Table \ref{tab:param}) and the same input ${\bf x}_{real}$. 
The third column results confirm that the MMC-GAN approach outperforms DNN-based classification.  This suggests the generalization ability of GAN-based methods independently of their type, although the best results were obtained with the CGAN structure.
\Update
Moreover, to assess the benefits of our multimodal approach, we repeated these experiments with the instruction sentence only I (${\bf x}_{real}={\bf x}_{instr}$), with the instruction and context sentences I+C (${\bf x}_{real}=\{{\bf x}_{instr}, {\bf x}_{c}\}$) and with vision input only V (${\bf x}_{real}={\bf x}_{v}$) in comparison with our multimodal approach I+C+V.

The results in the two first columns of Table \ref{tab:results} show that linguistic features are not enough to correctly predict the likelihood of a target area. The context sentence does not statistically significantly improve the results beccause the context is not informative enough. Nonetheless, this context is particularly useful for real experiments for guiding the robot's behavior. Similarly visual inputs does not give accurate results as reported in the third column. In contrast, the multi-modal gives accurate results that outperforms (by more than $20\%$) the other approaches. These results validate our method to disambiguate instructions.  \Update We can also emphasize that our model captures the correspondence between linguistic instruction and the scene. This is illustrated by the middle right sample in Fig. \ref{fig:samples} that is rejected since the visual input does not correspond to the instruction mentioning a drawer.\Done

The results reported for the GAN and CGAN networks are partial, since several experiments  were not able to converge ($5$ and $2$ for GAN, $4$ and $7$ for CGAN). Tweaking the network hyperparameters would certainly allow to improve the convergence rate. However, it is relevant to  emphasize this limitation of GAN-based approaches that has been studied in many recent works \cite{salimans2016improved, gulrajani2017improved}. In this case, WGAN can be an interesting alternative, since it offers a better stability but at a cost of slight accuracy decrease in our tests.  
\Done
Eventually, we evaluated our approach for the different  variations of $D$ and $G$. The best results, considering the full set of features, for each type of architecture are reported in Table \ref{tab:results2}. This time, these results emphasize that independently of their structures, GAN-based classifier outperforms DNN. The prediction results of MMC-GAN are improved by $4.8\%$ compared to the baseline, for the structure $4$.

\subsection{Error Analysis}\label{sec:error_analysis}
In this section we analyze the results categorization obtained by our MMC-GAN approach. The confusion matrix given in Table \ref{tab:conf} show the classification result for our best model ($87.5 \%$ of accuracy). It can immediately be noticed that most of the confusion in the categorization task are between classes $A4$ and $A3$ as well as between classes $A2$ and $A1$. One probable reason is that the difference in depth image for classes $A4$ and $A3$ (respectively for $A2$ and $A1$) is not sufficient because of the noise in the depth image. Our classifier mainly relies on the object depth value that when smeared with noise are not detected. Oppositely noise pixels could also be detected as object. For instance in the bottom right image in Fig. \ref{fig:samples}, noisy elements (black pixels) appear horizontally on the front side of the desk, which may explain the misclassifcation as $A2$ instead of $A1$. Furthermore, the bottom left image illustrates the case where the labeling strategy is not a strict process: this sample could also be labeled as $A3$ since there is some space appearing between the obstacles.

\begin{table}[t]
\normalsize
\caption{\small Maximum test set accuracy for the different network structures.}
\label{tab:results2}
\centering
\begin{tabular}{c c|c|c|c|c}
\hline
\multicolumn{1}{c}{}&\multicolumn{1}{c}{GAN}&\multicolumn{4}{|c}{Architecture}\\
 \cline{3-6}
 Method &  type& (1) & (2) & (3)& (4) \\
\hline
\hline
Baseline&-  &80.5&81.3&81.3& 83.4 \\
\hline
Ours& GAN &82.3 &83.3& 85.7& 87.0\\
\hline
Ours &CGAN &\bf 85.6 & \bf 85.7&\bf 86.7& \bf{87.5}\\
\hline
Ours&WGAN  &79.2&80.2&83.7&86.3 \\
\hline
\end{tabular}
\squeezeup

\end{table}

\begin{table}[h]
\normalsize
\caption{\small Confusion matrix of the test set for the best model.}\label{tab:conf}
\centering
\begin{tabular}{|c|c|cccc|}
\hline
\multicolumn{2}{|c}{}&\multicolumn{4}{|c|}{\bf Prediction }\\
\cline{3-6}
\multicolumn{1}{|c}{}&\multicolumn{1}{c|}{\bf $[\%]$}&$A1$ &$A2$ & $A3$ & $A4$ \\
\hline
&$A1$ &\cellcolor{gray!120}{\bf 94}& \cellcolor{gray!5}2&0& 4 \\

&$A2$  & \cellcolor{gray!10}4& \cellcolor{gray!80}{\bf 89} &\cellcolor{gray!10}3&\cellcolor{gray!10}4\\

 \rot{\rlap{\bf True}}&$A3$ & 0 & \cellcolor{gray!10}4 &\cellcolor{gray!78}{\bf 82} &\cellcolor{gray!40}14\\

&$A4$ &0&\cellcolor{gray!2}1&\cellcolor{gray!20}9 &\cellcolor{gray!90}{\bf 90}\\ 
\hline
\end{tabular}
\squeezeup
\squeezeup
\end{table}


Fortunately, for application on HSR, the robot would perform the task only for target areas classified as $A1$ or $A2$. Hence, the accuracy of our system drastically increases which leads to a potential  accuracy over $95 \%$. 

%% file: tex/conclusion.tex
\section{CONCLUSION}
Motivated by the increasing demand in domestic service robots, we proposed a method to understand ambiguous language instructions for carry-and-place tasks.  More explicitly, we proposed a method based on the instruction of the user and the scene situation  to predict target areas suitable for the task, by taking into account the space available and the robot's physical abilities.
The  following sums up our key contributions.
\begin{itemize}
 \item[$\bullet$ ]  We built a multimodal dataset  associating linguistic and visual inputs of target areas. 
 \item[$\bullet$ ]  We proposed the MultiModal Classifier GAN (MMC-GAN) which predicts the suitable target areas from multimodal inputs (linguistic and visual). Our results emphasize that MMC-GAN generalization ability outperforms the DNN-based classifier by  $4.1\%$ in accuracy.
  \item[$\bullet$ ] Several variations of MMC-GAN have been developed with the cutting edge methods derived from GAN: namely WGAN or CGAN. The best results were obtained from a CGAN architecture  with $87.5\%$ accuracy. 
\end{itemize}

 
In future work, we plan to extend MMC-GAN with a fully connected structure \Update(Extractor, Discriminator and Generator trained simultaneously) \Done which should be more convenient. A physical experimental study with non-expert users and HSR is also planned for a future work.  

%% file: tex/appendix.tex
\begin{appendices}

\section{GAN architectures}
\begin{table}[h]
\normalsize
\centering
\caption{\small Hidden layers structures and configurations for MMC-GAN }\label{tab:config}
\begin{tabular}{|c|c|l|}
\hline
 Num. &GAN & $100$, $100$, $100$, $100$ \\
\cline{2-3}
nodes  &CGAN  & $100$, $100$, $100$, $100$  \\
\cline{2-3}
 (G) &WGAN & $100$, $100$, $100$, $100$  \\
\hline
\hline
&(1) &  $100$, $200$ \\
\cline{2-3}
Num.&(2) & $100$, $200$, $400$ \\
\cline{2-3}
 nodes   & (3) &  $100$, $200$, $400$, $1000$ \\
\cline{2-3}
(D) &(4) &  $100$, $200$, $400$, $1000$, $1000$\\
\cline{2-3}
&(5)&  $100$, $200$, $600$, $400$, $1000$, $600$, $1000$ \\
\hline
\end{tabular}
\end{table}
\end{appendices}